\documentclass[10pt,journal,compsoc]{IEEEtran}
\usepackage{multirow}
\usepackage{bm}
\usepackage{comment}
\usepackage{booktabs}
\usepackage{algorithm,algpseudocode}
\usepackage{soul}
\usepackage{multirow}

\usepackage{amsmath,amsfonts,bm}

\def\eqref#1{equation~\ref{#1}}

\def\1{\bm{1}}

\def\vb{{\bm{b}}}

\def\ve{{\bm{e}}}

\def\vt{{\bm{t}}}

\def\vx{{\bm{x}}}

\DeclareMathAlphabet{\mathsfit}{\encodingdefault}{\sfdefault}{m}{sl}
\SetMathAlphabet{\mathsfit}{bold}{\encodingdefault}{\sfdefault}{bx}{n}
\newcommand{\tens}[1]{\bm{\mathsfit{#1}}}

\def\tB{{\tens{B}}}

\def\tM{{\tens{M}}}

\def\tW{{\tens{W}}}

\def\sX{{\mathbb{X}}}

\newcommand{\E}{\mathbb{E}}

\newcommand{\R}{\mathbb{R}}

\DeclareMathOperator*{\argmax}{arg\,max}
\DeclareMathOperator*{\argmin}{arg\,min}

\usepackage{booktabs} 
\usepackage{caption}
\usepackage{subcaption}
\usepackage{cleveref}
\usepackage{amsmath}

\usepackage[customcolors]{hf-tikz}

\ifCLASSOPTIONcompsoc
  \usepackage[nocompress]{cite}
\else
  \usepackage{cite}
\fi
\ifCLASSINFOpdf
\else
\fi
\begin{document}
%
\title{T-BFA: \underline{T}argeted \underline{B}it-\underline{F}lip Adversarial Weight \underline{A}ttack}
%
%
%
%

\author{Adnan Siraj Rakin, Zhezhi He, Jingtao Li, Fan Yao, Chaitali Chakrabarti and Deliang Fan}

%
%

\markboth{}%
{Shell \MakeLowercase{\textit{et al.}}: Bare Demo of IEEEtran.cls for Computer Society Journals}
%



\IEEEtitleabstractindextext{%
\begin{abstract}
Traditional Deep Neural Network (DNN) security is mostly related to the well-known adversarial input example attack. Recently, another dimension of adversarial attack, namely, attack on DNN weight parameters, has been shown to be very powerful. As a representative one, the Bit-Flip-based adversarial weight Attack (BFA) injects an extremely small amount of faults into weight parameters to hijack the executing DNN function. Prior works of BFA focus on \textit{un-targeted} attack that can hack all inputs into a random output class by flipping a very small number of weight bits stored in computer memory. This paper proposes the first work of \textit{targeted} BFA based (T-BFA) adversarial weight attack on DNNs, which can intentionally mislead selected inputs to a target output class. The objective is achieved by identifying the weight bits that are highly associated with classification of a targeted output through a \textit{class-dependent weight bit ranking} algorithm. Our proposed T-BFA performance is successfully demonstrated on multiple DNN architectures for image classification tasks. For example, by merely flipping 27 out of 88 million weight bits of ResNet-18, our T-BFA can misclassify all the images from 'Hen' class into 'Goose' class (i.e., 100\% attack success rate) in ImageNet dataset, while maintaining 59.35\% validation accuracy. Moreover, we successfully demonstrate our T-BFA attack in a real computer prototype system running DNN computation, with Ivy Bridge-based Intel i7 CPU and 8GB DDR3 memory.
\end{abstract}

}

\maketitle

\IEEEdisplaynontitleabstractindextext

%
\IEEEpeerreviewmaketitle

\IEEEraisesectionheading{\section{Introduction}\label{sec:introduction}}
\IEEEPARstart{I}{n} recent years, deep neural networks (DNNs) have achieved tremendous success in a wide variety of applications, including image classification~\cite{krizhevsky2010cifar,sun2019evolving}, speech recognition~\cite{hinton2012deep,haque2020experimental} and machine translation~\cite{luong2015deep,lu2020research}. Unfortunately, DNN models are not secure and the vulnerability of DNN models has been exposed by \cite{madry2018towards,goodfellow2014explaining} in their works on adversarial input example attack.

Recently, adversarial weight attacks have also been added to the security challenge of DNN models due to the security concern of model leakage and malicious fault injection into the computer system. First, the DNN model running in a computer is not secure. Many advanced computer side-channel attacks \cite{yan2020cache,xiang2020open,yudeepem,das2019x} have been shown to successfully extract DNN model parameters. Second, due to large size, DNN model integrity is difficult to guarantee in state-of-the-art performance-driven computing systems. In such systems, there are many different methods to inject a small amount of fault into the computing memory or path of DNN without alerting computing system. For example, several memory fault injection techniques, like Laser Beam Attack~\cite{roscian2013fault} or Row-Hammer Attack (RHA)~\cite{kim2014flipping,razavi2016flip}, can inject faults into a computer main memory (i.e., DRAM), causing severe threat to DNN computation in real computer. 

\begin{figure*}[ht]

    \centering
    \includegraphics[width=\textwidth]{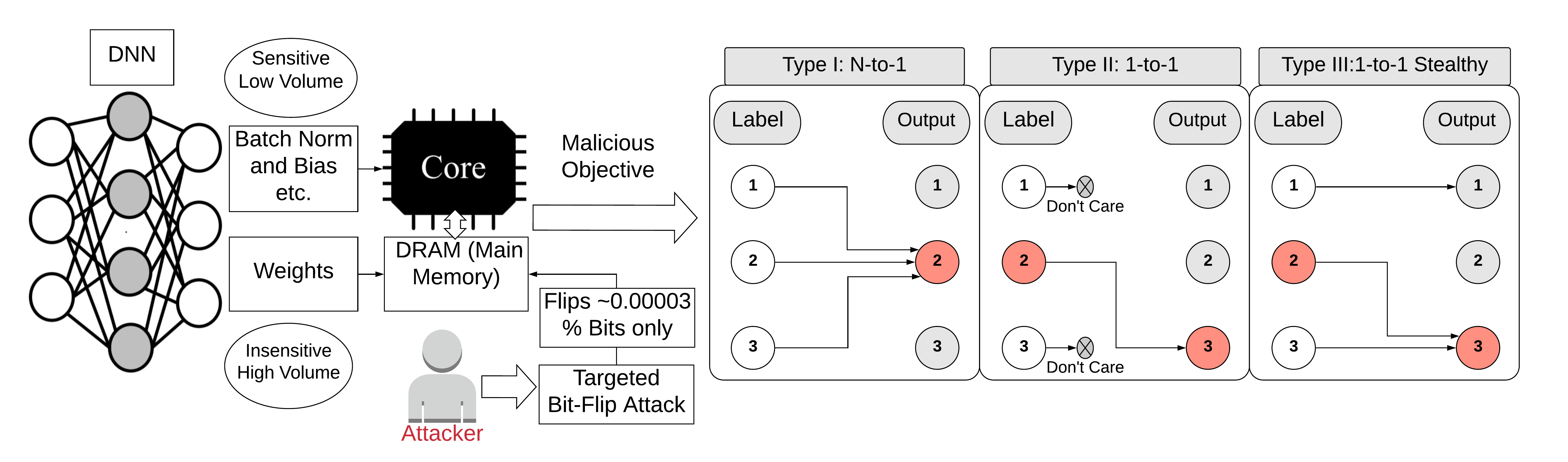} 
    \caption{Demonstration of Targeted Bit-Flip Attack (T-BFA) on the identified vulnerable bits achieving three distinct types of targeted attack objective.}
    \label{fig:RHA}
  
\end{figure*}

Due to the above existing vulnerability, as shown in the \cref{fig:RHA}, several recent works have leveraged such memory fault injection techniques to inject very minor faults (probably a few bits of error) into computer main memory (i.e. DRAM) to slightly modify the stored DNN model, successfully hijacking running DNN function \cite{rakin2019bit,liu2017fault,hong2019terminal,yao2020deephammer}.
From those works, a general conclusion is that weight quantized network is more robust than full precision (i.e. floating point number) version due to limited value range per weight, which is shown to require more bit flips (i.e. fault injection) in memory\cite{hong2019terminal,rakin2019bit}. 
Even so, the memory bit-flip based adversarial {un-targeted} weight attack (\textit{BFA}) proposed by our prior works in \cite{yao2020deephammer} and \cite{rakin2019bit} still experimentally demonstrate severe accuracy degradation of a fully-functional 8-bit quantized ResNet-18 on the ImageNet dataset to 0.1\% with only 13 bit-flips (out of 93 million bits), in real computer system. 
To summarize, BFA based adversarial weight attack is a real-world demonstrated and practical threat for DNN computing system. Comparing with adversarial input example attacks that require designing noise for each input separately, BFA based adversarial weight attack only needs to attack the model once to achieve the desired malicious behavior for all benign inputs.
 

Our prior \textit{un-targeted} bit-flip based adversarial weight attacks in \cite{yao2020deephammer,rakin2019bit} mainly focus on reducing the overall prediction accuracy to be as low as random guess. 
However, \textit{targeted} adversarial attacks~\cite{chen2017zoo,carlini2017towards} pose a greater threat for the following reasons: First, it gives the attacker precise control on the malicious objective and behavior. Second, a carefully crafted targeted adversarial attack objective can cause a devastating effect on the DNN output. For example, a targeted attack in self-driving car applications could cause a stop sign to be miss-classified as a high-speed limit sign, while keeping the accuracy of all other signs intact. It can be very stealthy and cause severe threat to DNN system security. 

However, all existing targeted attacks in the adversarial weight attack domain either fail to perform the attack effectively i.e., requiring a large number of weight modifications~\cite{zhao2019fault}, or target on a much more vulnerable full-precision DNN models~\cite{hong2019terminal,liu2017fault}.
Note that, a general conclusion is that a DNN with full precision weights is easier to attack. For example, as demonstrated in \cite{hong2019terminal}, a DNN can be forced to malfunction by just flipping the exponent bit, resulting in exponential change of the corresponding weight value. While, DNN with quantized weights is naturally noise resilient. More importantly, weight quantization is becoming a must-optimization for optimal efficiency and speed in many computing platforms, such as Google's TPU~\cite{jouppi2017datacenter}. Thus, in this work, we will only focus on attacking more robust quantized DNNs, rather than vulnerable full precision models.

This work is developed based on the submitting authors' prior conference proceedings \cite{rakin2019bit, yao2020deephammer}, which mainly introduce un-targeted bit-flip attack algorithm and its implementation in a real computer system, respectively. Compared with them, the novelty of this work lies in that we propose a new \textit{Targeted Bit-Flip Attack} (T-BFA) methodology. As far as we know, this is the first work of bit-flip based targeted adversarial weight attack on weight-quantized DNNs. We propose three variants of T-BFA as shown in the right panel of \cref{fig:RHA}: \textit{N-to-1} attack where inputs from $N$ source classes are hijacked to $1$ target class; \textit{1-to-1} attack where inputs from $1$ source class are mis-classified into $1$ target class; \textit{1-to-1 stealthy} attack where not only inputs from $1$ source class are hijacked to $1$ target class, but also the other class classification function remains the same. 

The novelty and contributions of this work are summarized as follows:

\begin{itemize}
    \item  Our proposed T-BFA is the first to demonstrate a successful targeted attack on noise resilient quantized DNNs through N-to-1 (I), 1-to-1 (II), and 1-to-1 stealthy (III) adversarial weight attacks by flipping a very small number of weight bits stored in computer memory. 
    

    \item To achieve the desired targeted attack objectives, we formulate three distinct loss functions associated with each type of attack. We leverage an iterative searching algorithm that can successfully minimize these loss functions to locate most vulnerable weight bits that are associated with a adversary target class.

    \item We evaluate T-BFA on a wide range of network architectures (e.g., ResNet, VGG and MobileNet-V2) for image classification using CIFAR-10 and ImageNet datasets. For example, the experiment on ResNet-18 using ImageNet dataset show that our proposed T-BFA can achieve 100\% attack success rate in miss-classifying all images in the `Hen' class into `Goose' class with only 27 bit-flips (out of 88 million bits) while keeping the test accuracy for other class images at 59.35\%.
    
    \item Finally, we demonstrate the practical feasibility of T-BFA in real computer attacks considering the adversary performs the attack by running an unprivileged user-space process on the machine (i.e., a strong adversary with system-level access permission is not required).
\end{itemize}

We organize the rest of the paper in the following manner: In section 2, we present background information of BFA and related target attack. In section 3, we present the targeted attack type and optimization method. Then we present the experimental setup and results in sections 4 and 5 respectively. It is followed by discussion in section 6 and conclusion in section 7.

\section{Background and Related Work}

\subsection{Adversarial Weight Attack} The recent developments in memory fault injection attacks~\cite{kim2014flipping,agoyan2010flip} have made it feasible to conduct an adversarial weight attack for a DNN model running in a computer. Among them, a row-hammer attack \cite{kim2014flipping} on Dynamic Random Access Memory (DRAM) is the most popular one since it can create a profile of memory bits stored inside the main memory (i.e., DRAM) and flip any bit of a given target address. The first few works that exploited row-hammer to attack DNN
 weights flipped the Most Significant Bits (MSB bits) of DNN parameters, such as the bias \cite{liu2017fault} or weight \cite{hong2019terminal}, and changed them to a significantly large value, thus degrading accuracy. However, those attacks were only targeted on a model with full precision (i.e. floating point) parameters and failed in more noise-resilient weight-quantized DNNs.

A major milestone in adversarial weight attack is the work in~\cite{rakin2019bit} which implemented a stronger version of BFA on an 8-bit fixed-point quantized network. The method in ~\cite{rakin2019bit} searches for the weight bits iteratively to gradually decrease DNN accuracy.
However, the BFA design in~\cite{rakin2019bit}
is an un-targeted attack. 
Even though it succeeds in hampering overall test accuracy, it fails to degrade the accuracy of a targeted class. Next, we briefly introduce the BFA attack threat model and methodology.

\subsection{Threat Model}

In this work, we follow the standard white-box attack threat model assumption, same as the previous bit-flip based adversarial weight attacks on quantized network ~\cite{rakin2019bit,rakin2019tbt, yao2020deephammer}. The previous BFA attack \cite{rakin2019bit} threat model. It assumes the attacker has access to model weights, gradients, and a portion of test data. Such a threat model is valid since previous works have demonstrated an attacker can effectively steal similar information (i.e., layer number, weight size, and parameters) through side-channel attacks~\cite{yan2020cache,xiang2020open,yudeepem,das2019x}. The attacker is denied access to any form of training information (i.e., training dataset, hyper-parameters). An attacker can only flip (0 to 1 or 1 to 0) identified bits in memory; no manipulation of input data is allowed.

\subsection{Weight Quantization \& Encoding}

In our evaluation of T-BFA, we adopt a similar weight quantization scheme as BFA attack \cite{rakin2019bit}. It is a layer-wise $N$-bits uniform quantizer for weight quantization. For each of the $l$-th layer, the quantization methodology can be described as:

\begin{equation}
\label{eqt:quan_stepsize}
    \Delta w_l = \textup{max}(\textbf{ W}_l^{r})/(2^{N-1}-1); \quad \textbf{W}_l^{\textup{r}} \in \R^d
\end{equation}

\begin{equation}
\label{eqt:quan_func}
    \textbf{W}_l = \textup{round}(\textbf{W}_l^{\textup{r}}/\Delta w_l) \cdot \Delta w_l
\end{equation}

where $d$ is the dimension of weight tensor, $\Delta w_l$ is the step size of weight quantizer, $\textbf{W}_l^{\textup{r}}$ is the full-precision weight of the corresponding quantized weight $\textbf{W}_l$. To circumvent the non-differential function (in \cref{eqt:quan_func}), popular straight-through estimator \cite{bengio2013estimating} is used to perform the training.

In our hardware evaluation, the computing system stores the signed integer in two's complement representation. Given one weight element $w \in \textbf{W}_l$, the conversion from its binary representation ($\vb=[b_{N-1},...,b_{0}
]\in \{0, 1\}^{N}$) in two's complement can be expressed as:
\begin{equation}
\label{eqt:twoscomplement}
w/\Delta w = bin(\vb) = -2^{N - 1}\cdot b_{N-1} + \sum_{i=0}^{N-2} 2^{i}\cdot b_{i}
\end{equation}
With the conversion relation described by $bin(\cdot)$ in \cref{eqt:twoscomplement}, we can inversely obtain the binary representation of weights $\textbf{B}$ (i.e. binary data stored in main memory) from its fixed-point counterpart as well.

\subsection{Un-Targeted BFA Attack Details}
Our preliminary Bit-Flip Attack (BFA) utilizes a combination of gradient ranking and progressive search to identify a set of vulnerable weight bits \cite{rakin2019bit,yao2020deephammer}. The objective of flipping the identified vulnerable weight bits is to degrade the overall test accuracy of the DNN. Thus at each iteration of the attack, the attacker will target at maximizing the inference loss function$\mathcal{L}$ w.r.t true label $\vt$ of a given test batch $\vx$: 
\begin{equation}
\label{eqt:BFA}
\begin{gathered}
\max_{\{\hat{\tB}_l^i\}}  ~\mathcal{L}\Big (f \big( \vx ; \{\hat{\tB}_l^i\}_{l=1}^{L} \big), \vt \Big) \\
\end{gathered}
\end{equation}
Here $l \in \{1, 2,...,L\}$ is the layer index, $\hat{\tB}_l^i$ is the bit representation of the weight matrix at the $i^{th}$ iteration at layer $l$ after flipping the bits in the original matrix $\tB_l$.

The progressive search of the BFA attack consists of two steps: i) \textit{in-layer search} and ii) \textit{cross-layer search}. Each of the step is performed progressively to identify vulnerable bits. First, for in-layer search, BFA will flip the top $n$ ranked bits (e.g., typically $n$=1) based on the gradient ($\argmax_{\tB_l} |\nabla_{\tB_l} \mathcal{L}|$) of every bit in each of the $l$ DNN layers. After flipping these bits at a given layer, the attacker evaluates the loss $\mathcal{L}$, and restores these flipped bits to the original state. Thus, a loss profile set $\{\mathcal{L}^{1}, \mathcal{L}^2, \cdot\cdot\cdot, \mathcal{L}^P\}$ is generated. For the cross-layer search step, the attacker identifies the attack layer with maximum loss: 

\begin{equation}
\begin{gathered}
 j = \argmax_l~\{\mathcal{L}^l\}_{l=1}^P
\end{gathered}
\end{equation}

Finally, the attacker goes to layer $j$ to perform the bit-flip in the $i^{th}$ iteration. To evaluate the attack efficiency, \cite{rakin2019bit} uses Hamming distance (i.e., effective bit-flips) between post attacked bits $\hat{\tB}_l^i$ and prior attack bits $\tB_l$ given by $\sum \mathcal{D}_{hd}(\hat{\tB}_l^i, {\tB}_l)$. In general, the optimization goal of un-targeted BFA is to lower the inference accuracy of DNN similar as random guess (i.e., 10 \% for CIFAR-10) with least number of bit-flips (i.e., $\min \sum \mathcal{D}_{hd}(\hat{\tB}_l^i, {\tB}_l)$). Even though BFA is a successful un-targeted adversarial weight attack on popular DNN architecture(i.e., ResNet, MobileNet-V2); it still fails to attack a particular target class. In the next subsection, we highlight some of the early attempts to conduct targeted weight attacks on DNN.

\subsection{Targeted Attack} In contrast to un-targeted BFA attack, a targeted attack has more precise control on the miss-classification behavior and can cause higher calamity. It is a well-investigated technique in adversarial input attack domain~\cite{carlini2017towards,chen2017zoo,madry2018towards}, where attacker finds additive input noise that decreases the loss function w.r.t a false target label for each input separately. 
Another form of targeted attack is the Trojan attack~\cite{gu2019badnets, Trojannn}, which typically requires hacking into the training supply chain for re-training network to force malicious behavior given a pre-designed input trigger\cite{gu2019badnets, Trojannn}. A recent more advanced Trojan attack also leverages BFA to inject Trojan by flipping close to one hundred bits with no need for network re-training or supply chain access\cite{rakin2019tbt}. However, it still needs the help of an input trigger, i.e. modifying inputs. This is out of the scope of this work, which limits injecting small errors to weight only. More closed related works are recent adversarial model parameter attacks that can perform a targeted attack without requiring an input trigger ~\cite{zhao2019fault,liu2017fault}, although they either require large amounts of weight perturbation or are evaluated only on the more vulnerable full precision model, not noise-resilient quantized model.



\section{Targeted Bit-Flip Adversarial Weight Attack}

\subsection{T-BFA Attack Objectives}
\label{sec:T-BFA_overview}

In this section, we present the proposed \textit{Targeted Bit-Flip adversarial weight Attack} (T-BFA) that results in mis-classification of benign inputs from their source category/categories (i.e., ground-truth) to the adversary target category, via a small number of malicious bit-flips on the quantized weight-bits of pre-trained DNN models. As depicted in~\cref{fig:RHA}, we propose three types of T-BFA with varying constraints, which are elaborated as follows:

\begin{itemize}

\item \underline{Type-I: N-to-1 Attack.} 
Given that the input data belong to one of $N$-classes, the objective of this T-BFA variant is to force the entire dataset $\mathbb{X} =\{\mathbb{X}_i\}_{i=1}^N$ with all $N$ classes (as source classes) to one adversary-selected target class. 
The objective function is formalized as:
\begin{align}
\min~\mathcal{L}_{\textrm{N-to-1}} = \min_{\{\tB\}}~\E_{\mathbb{X}} \mathcal{L}(f(\vx, \{\tB\}); \vt_q) 
\label{eqt:loss_T-BFA_I}
\end{align}
where $\{\tB\}$ is the quantized representation (in binary format) of weight tensor $\{\tW\}$ stored in computer memory. Given vectorized input $\vx \in \mathbb{X}$, $f(\vx,\{\tB\})$ computes quantized DNN inference output.
$\mathcal{L}(\cdot;\cdot)$ denotes the cross-entropy loss between DNN inference output and labels. 
$\vx$ and $\vt$ are input data and its corresponding ground-truth label.
For this attack, the ground-truth label term of source category\footnote{$\ve^{(i)}$ is the notation of one-hot code vector $[0,\dots,0,1,0,\dots,0]$ with a 1 at position $i$.} $\vt \in \ve^{(i)}, i\in \{1,...,N\}$ is tampered to the selected $q$-indexed target category $\vt_q\in \ve^{(q)}$. 

\item \underline{Type-II: 1-to-1 Attack.} In this T-BFA variant, adversary focuses on the mis-classification of input data $\mathbb{X}_p$ of single $p$-indexed source category into the $q$-indexed target category ($p\neq q$), without caring about the impact on the remaining categories $\mathbb{X}_{i\neq p}$. It can be modeled as:

\begin{equation}
\label{eqt:loss_T-BFA_II}
\min~\mathcal{L}_{\textrm{1-to-1}} = 
\min_{\{\tB\}}~\E_{\mathbb{X}_p} \mathcal{L}(f(\vx_p,\{\tB\});\vt_q); \quad \vx_p \in \mathbb{X}_p
\end{equation}

Type-II attack is a subset of Type I attack. But, such an objective is still practically useful to only attack a specific group or subset of inputs, where the type-I N-to-1 attack would flip many more un-necessary bits for all groups of inputs.

\item \underline{Type-III: 1-to-1 Stealthy Attack.}
In addition to the type-II 1-to-1 attack described above, this type-III attack is a stealthy version wit two objectives:  \textbf{1)} All the input data from $p$-indexed category $\mathbb{X}_p$ are classified into $q$-indexed target category, which is the same as~\cref{eqt:loss_T-BFA_II}; 
\textbf{2)} Meanwhile, it needs to maintain correct predictions of the input data excluded from the source category $\sX_j, j\in\{1,2,..,N\} \backslash \{p\}$.
This type-III attack could be achieved via the optimization of the two corresponding loss terms in the RHS of the following objective function:
\begin{align}
\label{eqt:loss_T-BFA_III}
\min~\mathcal{L}_{\textrm{1-to-1(S)}} 
= \min_{\{\tB\}}~\E_{\sX} \Big{(}\mathcal{L}(f(\vx,\{\tB\});\vt_q)\cdot \1_{\vx\in \sX_p} + \\ \mathcal{L}(f(\vx,\{\tB\});\vt) \cdot \1_{\vx\in \sX_j} \Big{)}
\nonumber
\end{align}

where $\1_{\textrm{condition}}$ returns 1 if the $\textrm{condition}$ is true, 0 otherwise.

\end{itemize}

\begin{figure*}[ht]

    \centering
    \includegraphics[width=0.8\textwidth]{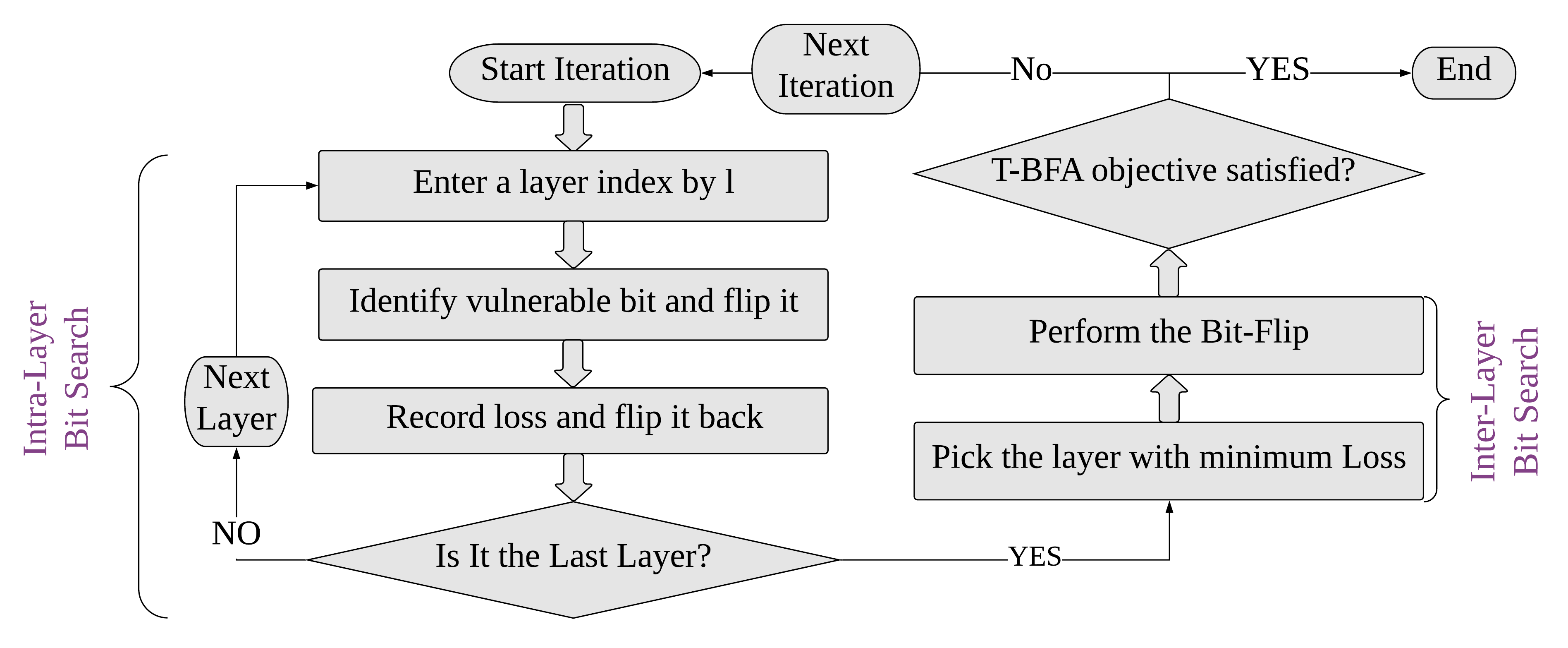} 
    \caption{Overview of T-BFA searching algorithm.}
    \label{fig:flow}
  
\end{figure*}

\begin{table*}[ht]

\centering
\caption{Test data splitting to conduct targeted attack from source class $t_\textrm{p}$ to target class $t_\textrm{q}$. CIFAR-10  data has 10k test images with each class containing 1000 test images and the ImageNet dataset has 50k test samples with each class containing 50 images. Note: ($t_\textrm{r}$) means images belong to any other class apart from the source class. }
\label{tab:data_dis}
\scalebox{0.9}{\begin{tabular}{ccccccc}
\toprule
Metrics & \begin{tabular}[c]{@{}c@{}} Attack \\ Batch Size\end{tabular} & \begin{tabular}[c]{@{}c@{}}\# of Data to \\  evaluate ASR ($X_\textrm{p}$)\\\end{tabular} & \begin{tabular}[c]{@{}c@{}}\# of Data to \\ evaluate Test acc. ($X_\textrm{r}$)\end{tabular} & \begin{tabular}[c]{@{}c@{}}Attack \\ Batch Size \end{tabular} & \begin{tabular}[c]{@{}c@{}}\# of Data to \\ evaluate ASR ($X_\textrm{r}$)\end{tabular} & \begin{tabular}[c]{@{}c@{}}\# of Data to \\evaluate Test acc. ($X_r$)\end{tabular} \\ \midrule
Dataset & \multicolumn{3}{c}{CIFAR-10} & \multicolumn{3}{c}{ImageNet} \\ 
\cmidrule(lr){2-4} \cmidrule(lr){5-7}
N-to-1  & 128 & 10k & 10k & 50 & 50k & 50k \\
1-to-1  & 500( $t_\textrm{p}$ ) & 500($t_\textrm{p}$) & 9k & 25($t_\textrm{p}$) & 25($t_\textrm{p}$) & 50k \\
1-to-1 (S) & 500($t_\textrm{p}$)+500($t_\textrm{r}$) & 500($t_\textrm{p}$) & 8.5k & 25($t_\textrm{p}$)+25 ($t_\textrm{r}$) & 25($t_\textrm{p}$) & 50k \\
\bottomrule

\end{tabular}}
\end{table*}




For a practical adversarial attack, to minimize attack effort, a critical constraint is to use limited number of malicious bit-flips on weight bits to achieve above defined attack objectives in~\cref{eqt:loss_T-BFA_I,eqt:loss_T-BFA_II,eqt:loss_T-BFA_III}. This could be modeled as a joint-optimization and represented by:
\begin{align}
\label{eqt:joint_optimization_1}
    \min~\mathcal{L}_{\textrm{T-BFA}}, ~
    \mathcal{L}_{\textrm{T-BFA}} \in \{\mathcal{L}_{\textrm{N-to-1}}, \mathcal{L}_{\textrm{1-to-1}}, \mathcal{L}_{\textrm{1-to-1(S)}}\};  \\
    \quad \textrm{s.t.}~\min_{\{\tB\}}~\mathcal{D}_{\textrm{hd}}( \{\hat{\tB}\}, \{\tB\}); \nonumber
\end{align}
where $\mathcal{D}_{\textrm{hd}}$ is the Hamming-distance between the weight-bit tensors of pre-attack model ($\{\tB\}$) and post-attack model($\{\hat{\tB}\}$). 
Instead of applying $\mathcal{D}_{\textrm{hd}}$ as an additional loss term in~\cref{eqt:loss_T-BFA_I,eqt:loss_T-BFA_II,eqt:loss_T-BFA_III} to form one combined multi-objective function, we follow the searching algorithm from our prior un-targeted BFA~\cite{rakin2019bit}, with several T-BFA-specific modifications; the details are given in the following subsection.  

\subsection{Vulnerable Weight Bits Searching Algorithm of T-BFA}

\label{sec:optimization_T-BFA}
The search of most vulnerable weight bits to be attacked by T-BFA can be generally described as an iterative process, wherein each iteration, only single weight-bit is identified followed by the malicious bit-flip. 
In the $k$-th iteration, the objective function~\cref{eqt:joint_optimization_1} is rephrased as:
\begin{equation}
\label{eqt:joint_optimization_2}
\min_{\{\tB^k\}}~\mathcal{L}_{\textrm{T-BFA}}; \quad \textrm{s.t.}~\mathcal{D}_{\textrm{hd}}(\{\tB^{k}\}, \{\tB^{k-1}\}) = 1
\end{equation}
where the single bit-flip is highlighted by defining inter-iteration Hamming distance $\mathcal{D}_{\textrm{hd}}$ as 1.
To minimize $\mathcal{L}_{\textrm{T-BFA}}$ with single bit-flip per iteration, we inherit and modify the progressive intra- and inter-layer bit search method in our prior un-targeted BFA scheme~\cite{rakin2019bit}.

Given a DNN model with $L$ layers (e.g., convolution layers), for one search iteration, the \textit{intra-layer bit search} is to identify one weight-bit per layer and traverse through all $L$ layers, thus returning $L$ weight-bit candidates. 
Then, the following \textit{inter-layer search} identifies one winner weight-bit out of $L$ weight-bit candidates brought up by the last step intra-layer search. This identified winner weight-bit will be flipped and the search process goes to the next iteration. The whole progressive search process ends when the adversary defined attack objective is achieved as shown in \cref{fig:flow}. We will describe such two-step progressive searching method in one iteration-$k$ in the following paragraphs.

\textbf{Intra-layer Bit Search.}
For layer indexed by $l$, the intra-layer bit search identifies one(or more) weight-bit candidate(s) w.r.t two criteria: 
1) identifying the weight-bit with the highest gradient; 
2) flipping along the direction of bit-gradient. Note that, in our prior un-targeted BFA~\cite{rakin2019bit}, the weight-bit is flipped along the opposite direction of bit-gradient, as it performs loss maximization, instead of minimization defined in \cref{eqt:joint_optimization_2} in this work. To perform the bit-flip, we adopt the same mask technique in \cite{rakin2019bit} to check whether the chosen bit can be flipped in the desired direction. These two criteria can be mathematically described as:

\begin{align}
\label{eqt:bit_ranking}
\argmax_{\tM_l^k, b_l^k} | \nabla_{\tB_l^{k-1}}  \mathcal{L}_{\textrm{T-BFA}}^k  |;
\quad \textrm{s.t.} ~ b_l^k = \\ \textrm{clamp}\Big{(} b_l^{k-1} - \textrm{sign}( \nabla_{b_l^{k-1}} \mathcal{L}_{\textrm{T-BFA}}^k )\Big{)}, ~ b_l^k \neq b_l^{k-1} \nonumber
\end{align}

where $\tM_l^k$ is the mask that indicates the location of the identified bit within weight-bit tensor $\tB_l^{k-1}$ and its value $b_l^k\in\{0,1\}$. $\textrm{clamp}(\cdot)$ is the clamping function with 0 and 1 as lower and upper bound. The intra-layer bit search traverses through all the layers to generate the weight-bit candidate set, $\{ \tM_l^k\}_{l=1}^{L}$. Meanwhile, for each weight-bit candidate in~$\{ \tM_l^k\}_{l=1}^{L}$, the corresponding T-BFA loss is profiled $\{\mathcal{L}_{\textrm{T-BFA},l}^k\}_{l=1}^L$ after the identified weight-bit is flipped.

\textbf{Inter-layer Bit Search.} Based on the intra-layer search outcomes (i.e., $\{ \tM_l^k\}_{l=1}^{L}$), the inter-layer search performs straight-forward comparison to identify the winner weight-bit candidate with minimal profiled loss as the weight-bit to attack in iteration-$k$. This process can be expressed as follows:
\begin{equation}
\label{eqt:inter_layer}
    \argmin~\{\mathcal{L}_{\textrm{T-BFA},l}^k\}_{l=1}^L
\end{equation}
When the winner weight-bit is identified, it will be flipped to perturb the DNN model with only one-bit difference with the model in the previous iteration. Then, another new search iteration will start with this new model parameters. The whole process ends when the attack goal is achieved.  

\begin{table*}[ht]

\centering
\caption{N-to-1 Attack: number of bit-flips (mean$\pm$std) required to classify all the input images to a corresponding target class with 100\% ASR. In each case, test accuracy drops to 10\%.}
\label{tab:t1}
\scalebox{0.9}{
\begin{tabular}{@{}cccccccccccc@{}}
\toprule
Class & 0 & 1 & 2 & 3 & 4 & 5 & 6 & 7 & 8 & 9& Average \\ \midrule
ResNet-20 & 4.0 $\pm$ 0 & 4.6 $\pm$ 0.9 & 5.0 $\pm$ 2.2 & 6.2 $\pm$ 2.3 & 4.6 $\pm$ 0.9 & 5.2 $\pm$ 1.6 & 6.8 $\pm$ 1.9 & 4.4 $\pm$ 1.7 & 5 $\pm$ 2.2 & 4.8 $\pm$ 1.8& 5.1 \\
VGG-11 &3.0 $\pm$ 0.0 & 3.0 $\pm$ 0.0 & 3.0 $\pm$ 0.0 & 3.0 $\pm$ 0.0 & 2.8 $\pm$ 0.4 & 2.0 $\pm$ 0.0 & 3.0 $\pm$ 0.0 & 3.2 $\pm$ 0.4 & 3.0 $\pm$ 0.0 & 3.0 $\pm$ 0.0 & 3.0\\  \bottomrule
\end{tabular}}

\end{table*}

\begin{figure*}[t]
    \centering
    \includegraphics[width=0.95\textwidth]{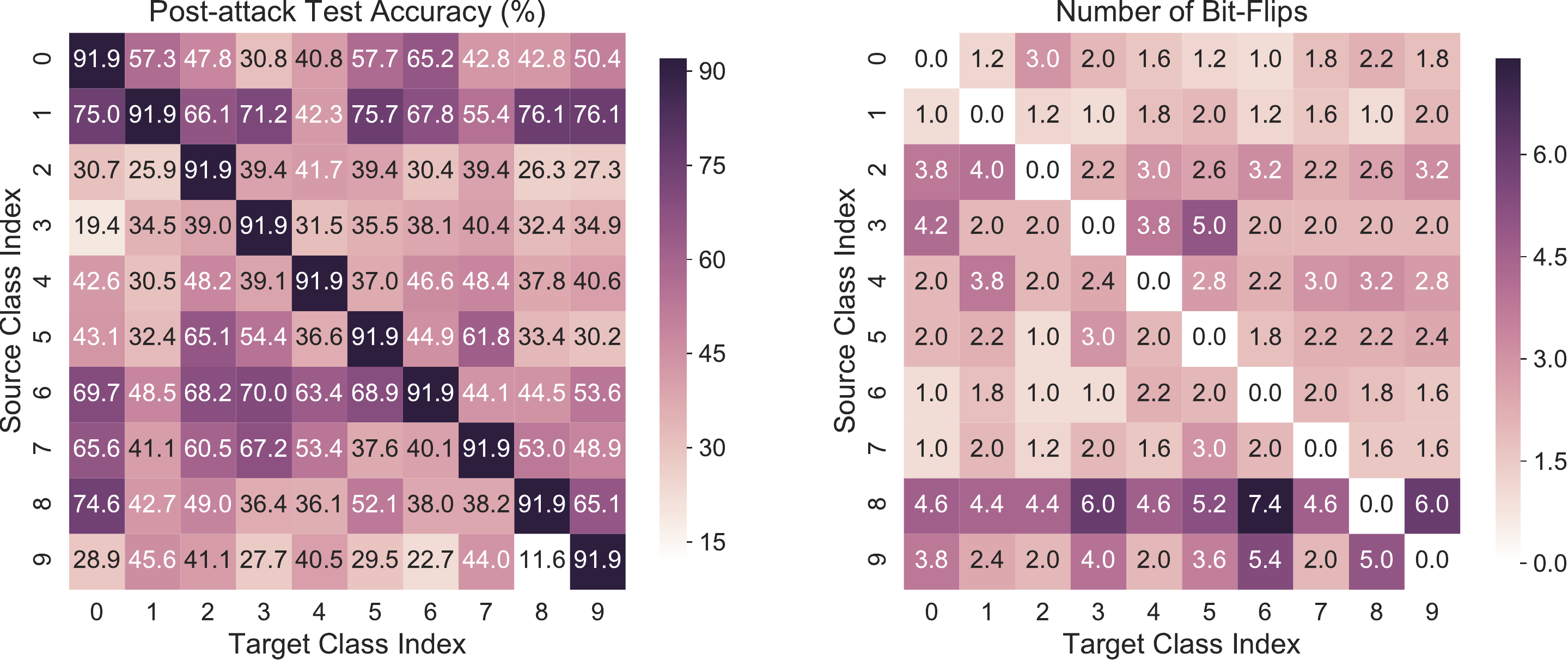} 
    \caption{Type II: 1-to-1 attack on ResNet-20 between source class and target class. The left subplot shows post attack test accuracy and the right subplot shows average number of bit-flips required for the attack. 
    }
    \label{fig:tar2}
 
\end{figure*}

\begin{figure*}[t]
    \centering
    \includegraphics[width=1\textwidth]{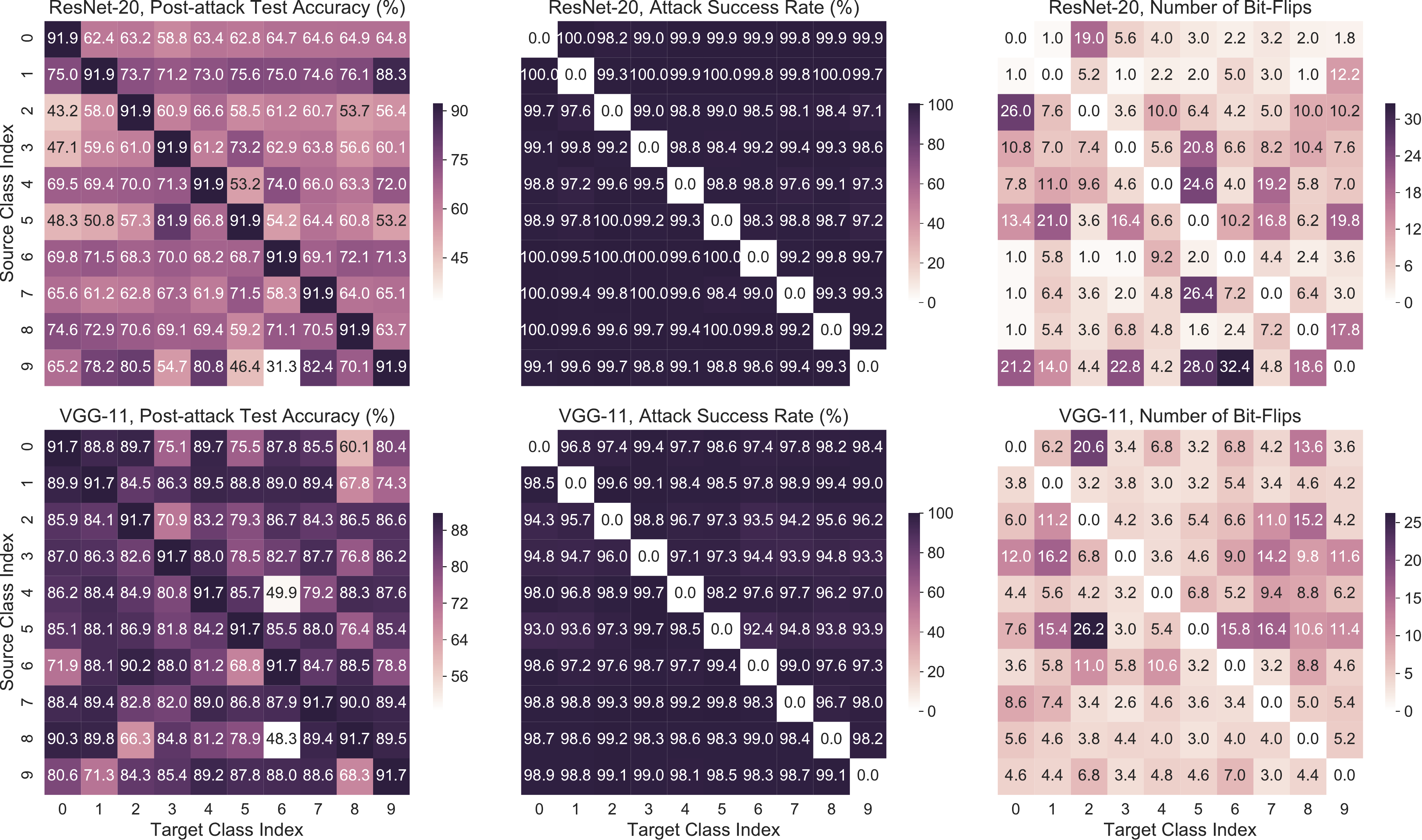} 
    \caption{Type III: 1-to-1 (S) attack post attack test accuracy, attack success rate and avg. \# of bit-flips for five rounds of attacks for both Resnet-20 and VGG-11 Networks. 
   }
    \label{fig:tar3}

\end{figure*}

\section{Experimental Setup}

\subsection{Dataset}

In our experiment, we test our T-BFA in image classification using two popular datasets i) CIFAR-10 \cite{krizhevsky2010cifar} and ii) ImageNet \cite{krizhevsky2012imagenet}. CIFAR-10 is a popular visual recognition dataset, which includes 60k images combined with training and test set. Each RGB image has a size of $32\times32$ evenly sampled from 10 categories. The data augmentation technique is identical to previous methods \cite{he2016deep}. ImageNet is a large dataset containing 1.2M training images. The size of the images of the ImageNet dataset is 224 $\times$ 224 that is equally divided into 1000 distinct classes. 

\subsection{Dataset configuration for attack}

 In~\cref{tab:data_dis}, we provide an overview of data division to conduct each type of attack.
To conduct an N-to-1 attack on CIFAR-10 and ImageNet, we randomly choose a test batch from the test dataset. However, to evaluate 1-to-1 or 1-to-1(S) attack, we require a subset of source class ($t_\textrm{p}$) test images. Since the CIFAR-10 dataset has 1k images in each class, we use 500 images to perform the attack and the remaining 500 images for evaluating the \textit{Attack Success Rate (ASR)}.
Since the ImageNet dataset has only 50 images per class, we conduct the attack using 25 images from the source class and use the remaining 25 images for evaluating ASR. Similar to BFA \cite{rakin2019bit} attack, we observe the effect of attack batch size plays a minor role in the attack performance. Furthermore, for ImageNet, we always evaluate test accuracy on the whole test dataset of 50k images because the amount of test data used to perform the attack (e.g., 50) is negligible compared to 50k test images. The mean and standard deviation numbers are calculated over 5 trial runs for CIFAR-10 and 3 trial runs for ImageNet. Also, we terminate attacks when the ASR reaches higher than 99.99\% or remains the same for three successive iterations. 

\subsection{DNN Architectures}
For CIFAR-10 dataset, we evaluate the attack against popular ResNet-20 \cite{he2016deep} and VGG-11 \cite{simonyan2014very} networks. We use the same pre-trained models with exact configuration as \cite{he2020defending}. For ImageNet results, we evaluate our attack performance on MobileNetV2 \cite{sandler2018mobilenetv2}, ResNet-18 and ResNet-34 \cite{he2016deep} architectures. For each of the model, we directly download a pre-trained model from PyTorch Torchvision models \footnote{https://pytorch.org/docs/stable/torchvision/models.html} and perform an 8-bit post quantization and encoding as described in previous section 2.3. 

\subsection{T-BFA Attack Setup in a Real Computer}
To demonstrate T-BFA attack on a DNN running in a real computer, we implement a DRAM fault injection method using the same computer system setup as our prior work in~\cite{yao2020deephammer}. Here the adversary performs the row-hammer attack to inject bit flips in DRAM by running an unprivileged user-space process on the machine (i.e., a strong adversary with system-level access permission is \emph{not} required). Our attack is evaluated on a computer with Intel Ivy Bridge-based processor and dual-channel 8GB DDR3 memory with two DIMMs. Each DRAM DIMM has 16 banks and Each each bank has 32768 rows.  
We implement a double-sided row-hammer attack where the attacker controls two neighboring rows of a victim row (i.e., rows that store DNN weights) to induce bit flip in the victim DNN model. To achieve such memory layout, we reverse-engineer the DRAM addressing scheme using the technique demonstrated in~\cite{pessl2016drama}. 

We first perform memory templating that scans DRAM rows to collect information about flippable bits (i.e., bit flip profile) in the main memory. Such an off-line DRAM profiling can be done in isolation in the attacker's memory space, and thus does not corrupt or crash the system \cite{razavi2016flip,roscian2013fault}. We employ the stripe data pattern (1-0-1 and 0-1-0) with a double-sided row-hammer in order to extract most bit flips \cite{yao2020deephammer,kim2014flipping}. The bit-flip profile keeps track of locations and flip directions for vulnerable memory cells. 

After our T-BFA algorithm search is finished, the attacker generates a set of bit offsets in the target DNN's weight file. 
The weight parameters in the weight files are organized as physical pages (typically in the size of 4KB) in DRAM. To ensure that these identified target bits could be flipped, the attacker needs to ensure that DRAM pages holding the targeted weight parameters are located in the desirable DRAM rows. Particularly, the attacker manipulates the Operating System through page cache to \emph{massage} the memory~\cite{gruss2018another,{gruss2018another}} so that the target weight bits are stored in flippable DRAM cells with the right flip direction (i.e., either 1$\rightarrow$0 or 0$\rightarrow$1). The attacker then performs double-sided row-hammering through frequently accessing its own data (the neighboring rows) to incur sufficient disturbance to DNN's memory row to achieve the targeted bit flips. 
In some cases, if the identified bits are not flippable in hardware, a new set of vulnerable bit candidates from our T-BFA algorithm will be generated by freezing the previous set (e.g., in case that the bit flip found in the profile can not be repeated at runtime). This ensures our software algorithm runs independently of system attacks. In our real computer attack experiments, we successfully validate all types of T-BFA on different DNN architectures in our real computer attack as will be reported later.

\subsection{Evaluation Metrics}
Two metrics are used in this work for attack evaluation:
\textit{Post-attack test accuracy (TA\%)} and \textit{Attack Success Rate (ASR\%)}.

\textbf{Post-Attack Test Accuracy (TA\%):} The Post-attack test accuracy is the inference accuracy of the post-attack model on test set. To evaluate the test accuracy after the attack, we only use a portion of the test data ($X_\textrm{r}$ in \cref{tab:data_dis}) which does not contain any image from the source class; since all the source class images will be miss-classified to the target class after the attack.

\textbf{Attack Success Rate (ASR\%):} The ASR is the percentage of source class images(i.e., $X_\textrm{p}$ in \cref{tab:data_dis}) successfully classified into the adversary target class via T-BFA. To evaluate ASR, we only use $X_\textrm{p}$ portion of source class data shown in table \ref{tab:data_dis}. The attacker does not use this portion of the source class images during the attack for 1-to-1 and 1-to-1 (S). However, for N-to-1 (S) $X_\textrm{p}$ contains the whole test dataset, since, by definition, the attack should classify all the test images into one target class.

\label{sec:exp}



\begin{table*}[ht]

\centering
\caption{Performance of T-BFA variants on ImageNet (from Hen class (i.e., label \emph{8}) to Goose class (i.e., label \emph{99})). The original test accuracies of ResNet-18, ResNet-34 and MobileNet-V2  are 69.23\%, 75.5\% and 72.01\%, respectively.}
\scalebox{0.95}{
\begin{tabular}{@{}|c|ccc|ccc|ccc|@{}}
\toprule
Type & \begin{tabular}[c]{@{}c@{}}Attack \\ Success\\ Rate (\%)\end{tabular} & \begin{tabular}[c]{@{}c@{}}Test\\ Accuracy\\ (\%)\end{tabular} & \begin{tabular}[c]{@{}c@{}}\# of \\ Bit-Flips\end{tabular} & \begin{tabular}[c]{@{}c@{}}Attack \\ Success\\ Rate (\%)\end{tabular} & \begin{tabular}[c]{@{}c@{}}Test\\ Accuracy\\ (\%)\end{tabular} & \begin{tabular}[c]{@{}c@{}}\# of \\ Bit-Flips\end{tabular} & \begin{tabular}[c]{@{}c@{}}Attack \\ Success\\ Rate (\%)\end{tabular} & \begin{tabular}[c]{@{}c@{}}Test\\ Accuracy\\ (\%)\end{tabular} & \begin{tabular}[c]{@{}c@{}}\# of \\ Bit-Flips\end{tabular} \\ \midrule
N-to-1  & 99.78 $\pm$ 0.27 & 0.23 $\pm$ 0.18 & 32.6 $\pm$ 8.2    & 99.99 $\pm$ 0& 0.1 $\pm$ 0 & 21 $\pm$ 4 & 100 $\pm$0 & 0.1 $\pm$ 0  & 17.3 $\pm$ 3.29 \\
1-to-1  & 100 $\pm$ 0 & 32.13 $\pm$ 14.4 & 16.7 $\pm$ 1.24  & 100 $\pm$ 0 & 23.74 $\pm$ 1.71 & 9.33 $\pm$ 0.94 & 100 $\pm$ 0 & 1.19 $\pm$ 0.22 & 13 $\pm 1.41$ \\
1-to-1 (S) &   100 $\pm$ 0 & 59.48 $\pm$ 2.9 & 27.3 $\pm$ 16.7 & 100 $\pm$ 0&  58.33 $\pm$ 3.29 & 40.33 $\pm$ 30.32 & 98.67 $\pm$ 1.89  & 33.99 $\pm$ 4.93 & 45.33 $\pm$ 21.74  \\
 & \multicolumn{3}{c|}{ResNet-18 (\# of parameters: 11M)} & \multicolumn{3}{c|}{ResNet-34 (\# of parameters: 21M)} & \multicolumn{3}{c|}{MobileNet-V2 (\# of parameters: 2.1M)} \\ \bottomrule

\end{tabular}}

\label{tab:ima}

\end{table*}

\begin{table*}[ht]

\centering
\caption{Comparison with Competing Methods We directly report the numbers from the respective papers for~\cite{liu2017fault,zhao2019fault}. For~\cite{rakin2019tbt,rakin2019bit} we run the attack on ResNet-20 8-bit quantized network.}
\label{tab:cmp}
\scalebox{0.9}{
\begin{tabular}{cccccc}
\toprule
Method & \begin{tabular}[c]{@{}c@{}}\# of Data used to \\ evaluate ASR \end{tabular} & \begin{tabular}[c]{@{}c@{}} ASR (\%)\end{tabular} & \begin{tabular}[c]{@{}c@{}} Post Attack \\ Test Accuracy (\%)\end{tabular} & \begin{tabular}[c]{@{}c@{}}\# of \\ Bit-Flips\end{tabular} & \begin{tabular}[c]{@{}c@{}}Model\\ Precision\end{tabular} \\ \midrule
Untargeted-BFA (I) \cite{rakin2019bit} &10k &- & 10.27 & 28& 8-bit\\
\textbf{Proposed N-to-1(I)} &10k &100 & 10 & \textbf{4} & 8-bit  \\
\midrule
SBA (II) \cite{liu2017fault}& 100 &100 & 60.0 & 1& full-precision \\
\textbf{Proposed 1-to-1 (II)} & \textbf{1000} &100 & 10 & 3.2& \textbf{8-bit} \\ \midrule
TBT (III) \cite{rakin2019tbt} & 10k &93.89 & 82.03 & 199 & 8-bit\\
GDA (III) \cite{liu2017fault}& 100&100 & 81.66 & 198& full-precision \\
Fault Sneaking (III) \cite{zhao2019fault} &16&100 & 76.4 & \textgreater{2565}& full-precision\\
\textbf{Proposed 1-to-1 (s) III} &1000 &99.3 & \textbf{88.3} & \textbf{12.2} & \textbf{8-bit}\\ \bottomrule
\end{tabular}}

\end{table*}

\section{Experimental Results}
\subsection{Experiments on CIFAR-10}

\textbf{N-to-1 Attack.}
For CIFAR-10, the proposed N-to-1 attack can successfully reach 100\% ASR for both VGG-11 and ResNet-20 architectures on each target class. As shown in~\cref{tab:t1}, the range of average bit-flips required to achieve 100\% ASR is between $4\sim6.8$ and $2.8\sim3$ for ResNet-20 and VGG-11, respectively. So for the N-to-1 attack, VGG-11 requires a consistently fewer number of bit-flips than ResNet-20 for all CIFAR-10 classes.

\underline{\textbf{\textit{Take-Away 1.}}}
Our analysis of the N-to-1 attack shows that there is no particular target class that is easier or more difficult to attack. Thus we conclude that the input feature patterns play a small role in resisting the attack, while the network architecture plays a more important role.

\textbf{1-to-1 Attack.}
In this version of T-BFA, the attacker performs 1-to-1 miss-classification with fewer number of bit-flips (see \cref{fig:tar2}) in comparison to the N-to-1 version (see \cref{tab:t1}). For most of entries shown in~\cref{fig:tar2}, the 1-to-1 attack requires only 1-2 bit-flips to achieve 100\%ASR with a few exceptions. 
Overall, for all possible combinations of classes, T-BFA successfully achieves 100\% 1-to-1 miss-classification with a range of $1\sim7.4$ bit-flips. 

\underline{\textbf{\textit{Take-Away 2.}}} 1-to-1 attack requires, in general, less bit-flips compared to N-to-1 attack. This is expected since mis-classifying all N classes are more difficult than mis-classifying just one class.

\textbf{1-to-1 Stealthy (S) Attack.}
Our evaluation of 8-bit quantized ResNet-20 and VGG-11 models shows a 91.9\% and 91.6\% baseline CIFAR-10 test accuracy, respectively. As shown in~\cref{fig:tar3}, after attack, the accuracy of ResNet-20 has a larger drop. The average test accuracy after five rounds of attack is between $31.3\sim88.3$\% for ResNet-20. On the other hand, VGG-11 maintains a better test accuracy with a range of $48.3\sim 90.1\%$. 

Our T-BFA is effective in attacking ResNet-20 network by achieving ASR higher than 97\% for all combinations of source and target classes. However, VGG-11 shows slightly better resistance to the attack with an ASR range of 93-99\% for different combinations. This is consistent with prior work which also shows that denser networks (i.e., VGG-11, VGG-16) have better resistent to both adversarial weight attack~\cite{rakin2019tbt} and input attack~\cite{madry2018towards}. 
While for both networks, some classes are more vulnerable than others, most source class and target class combinations require less than 10 bit-flips to conduct the 1-to-1 stealthy attack.

\underline{\textbf{\textit{Take-Away 3.}}} A compact network, like ResNet-20 with 0.27M parameters, has less capacity to learn the dual objective function in 1-to-1 (S) attack through a small number of bit-flips in comparison to denser network, like VGG-11 with 132M parameters. As a result, the test accuracy drop for a compact network, like ResNet-20, is higher.

\subsection{Experiments on ImageNet}

ImageNet dataset has a much larger number of output classes compared to CIFAR-10. We do not have the space to report all targeted attack results, thus we randomly pick one combination of target attack  (Hen class -> Goose class) to show our method's efficiency. For N-to-1 attack, \cref{tab:ima} shows that T-BFA requires 32, 21 and 17.3 bit-flips, on average, for ResNet-18, ResNet-34 and MobileNet-V2, respectively. Aligning with the observation for CIFAR10, it can be seen that a more compact network is more vulnerable to the N-to-1 attack. For 1-to-1 (S) attack, a compact network, e.g., MobileNet-V2 (with 2.1M parameters), fails to maintain a reasonable test accuracy (i.e., 33.9\%). While larger networks, such as ResNet-18 and ResNet-34, could maintain a reasonable test accuracy (i.e., $\sim$ 59\%) while achieving 100\% ASR. Those experiments results also align with our observation for CIFAR10.

\underline{\textbf{\textit{Take-Away 4.}}} In the case of ImageNet dataset, a large number of output classes increase the attack difficulty for T-BFA. However, in consistent with CIFAR-10 observations, it is easier to conduct 1-to-1 (S) attack on a network with higher capacity. The  larger optimization space helps achieve dual objectives of maintaining reasonable test accuracy as well as achieving very high ASR.

\begin{table*}[ht]
\centering
\caption{T-BFA performance against existing BFA defense techniques. PTA indicates Post-attack test accuracy.
}
\label{tab:CVPR20}
\scalebox{0.95}{
\begin{tabular}{@{}ccccccccccc@{}}
\toprule
\textbf{Class} & \begin{tabular}[c]{@{}c@{}}Clean\\ Model\end{tabular} & \multicolumn{3}{c}{\textbf{N-to-1}} & \multicolumn{3}{c}{\textbf{1-to-1}} & \multicolumn{3}{c}{\textbf{1-to-1(S)}} \\ \midrule
 & TA(\%) & PTA(\%) & ASR (\%) & \begin{tabular}[c]{@{}c@{}}\# of \\ flips\end{tabular} & PTA (\%) & ASR(\%) & \begin{tabular}[c]{@{}c@{}}\# of \\ flips\end{tabular} & PTA(\%) & ASR(\%) & \begin{tabular}[c]{@{}c@{}}\# of \\ flips\end{tabular} \\ \midrule
\begin{tabular}[c]{@{}c@{}}8-Bit \end{tabular} & 91.9 & 10.0 & 100.0 & 5.2 & 49.0 & 100.0 & 4.4 & 66.3 & 99.2 & 3.6 \\
8-Bit (PC) \cite{he2020defending}& 91.29 & 10.0 & 100.0 & 5.6 & 47.7 & 100.0 & 3.0 & 66.49 & 97.84 & 8.6 \\
Binary \cite{he2020defending} & 88.24 & 10.0 & 100.0 & 35.5 & 61.94 & 100.0 & 17 & 72.98 & 98.4 & 16 \\ \bottomrule
\end{tabular}}

\end{table*}

\subsection{Comparison with Other Competing Methods}

In this section, we compare our proposed T-BFA with most recent existing works of targeted attacks \cite{rakin2019bit,rakin2019tbt,liu2017fault,zhao2019fault} in adversarial weight attack domain. 

As shown in \cref{tab:cmp}, our proposed N-to-1 targeted attack achieves the same objective as \cite{rakin2019bit} with \textbf{7} $\times$ less number of bit-flips. Moreover, unlike un-targeted BFA (i.e., randomly classifying all inputs to a random class), our N-to-1 attack has precise control on the target output class.
Other stronger versions of previous targeted attacks, such as GDA \cite{liu2017fault} and fault sneaking attacks \cite{zhao2019fault}, have shown superior results (100\% ASR) against a weaker threat model (i.e., full-precision model or the attack is evaluated against only 100 images). However, our proposed T-BFA 1-to-1 (s) outperforms both \cite{liu2017fault},\cite{zhao2019fault} on a quantized network with \textbf{16} $\times$ and \textbf{210} $\times$ fewer number of bit-flips. Among the Neural Trojan works \cite{Trojannn,gu2019badnets,rakin2019tbt}, Targeted Bit Trojan (TBT) follows a more strict threat model and performs the attack with the least number of bit-flips. However, our 1-to-1 (s) proves to be much more efficient than TBT w/o input trigger, as required in the targeted Trojan attack, like TBT. In comparison, T-BFA achieves a higher test accuracy and higher ASR with \textbf{16} $\times$ fewer bit-flips.

The analysis presented above for both BFA \cite{rakin2019bit} and TBT \cite{rakin2019tbt} uses the same architecture, data-set, and number of attack samples as ours. It shows T-BFA outperforms these two powerful attacks fairly in terms of all three evaluation metrics: \# of flips, attack success rate, and post-attack accuracy. However, for attacks from \cite{liu2017fault} and \cite{zhao2019fault} with no open-source code, we report the attack performance directly from their respective papers, where their attack methods are evaluated using a more vulnerable full-precision DNN model, rather than weight-quantized model. Even so, our method still outperforms theirs in terms of all above three metrics on a more robust quantized DNN. 
These comparison still serves the purpose to show the strength of our method, mainly because all the previous adversarial weight attacks \cite{hong2019terminal,rakin2019bit} have concluded that attacking a quantized network (8-Bit) is much more difficult than attacking a full-precision model. 

\subsection{Attacking Real Computer Running DNNs:}

In a real computer main memory, an 8-bit quantized DNN with $M$ number of weights contains ($M/4096$) physical memory pages (4KB) and within each page, one bit has an offset range (0-32767). We evaluate all three types of T-BFA on our prototype computer hardware (described earlier) running ResNet-20, VGG-11, and MobileNet-V2 summerized in \cref{tab:sum}. 

\begin{table}[ht]
\centering
\caption{T-BFA Attack on DNNs Running in a Real Computer}
\label{tab:sum}
\scalebox{0.9}{
\begin{tabular}{@{}ccccc@{}}
\toprule
Network & \begin{tabular}[c]{@{}c@{}}Attack\\ Type\end{tabular} & \begin{tabular}[c]{@{}c@{}}ASR\\ (\%)\end{tabular} & \begin{tabular}[c]{@{}c@{}}Post\\ Attack\\ Accuracy\\ (\%)\end{tabular} & \begin{tabular}[c]{@{}c@{}}Number \\ Of Flips\end{tabular} \\ \midrule
ResNet-20 (CIFAR-10) & I & 88.92 & 19.88 & 2 \\
MobileNet-V2 (ImageNet) & II & 96.8 & 2.2 & 11 \\
VGG-11 (CIFAR-10) & III & 98.6 & 80.6 & 2 \\ \bottomrule
\end{tabular}}
\end{table}

In our real computer attack system, by flipping ResNet-20 bit locations: (page \# 65 offset \# 12113);(page \# 1 offset \# 12600), attacker can achieve 88.92 \% ASR on \textbf{Type I} attack (e.g., class 2). Similarly, by flipping two bits of VGG-11:  (page \# 2379 offset \# 21352) ; (page \# 2378 offset \# 20504), attacker achieves 98.6\% ASR for \textbf{type III} (class 9 $\rightarrow$ 1) on CIFAR-10. We test ImageNet results on MobileNet-V2 for \textbf{type II} attack (class 8 $\rightarrow$ 99). By flipping these 11 bit locations: (page \# 1 offset \# 7392) ; (page \# 131 offset \# 12883); (page \# 3 offset \# 25971); (page \# 114 offset \# 22842);(page \# 143 offset \# 10335);(page \# 281 offset \# 16537);(page \# 298 offset \# 3298);(page \# 304 offset \# 21736);(page \# 285 offset \# 14549);(page \# 143 offset \# 9359);(page \# 465 offset \# 19993), attacker would achieve 96.8 \% ASR. Note that, due to the consideration of bit flip profile, the targeted bits can be flipped successfully in our physical testbed (\emph{the online row-hammer exploitation takes less than 30 seconds}).
\begin{table*}[ht]

\centering
\caption{Result of varying the quantization bit-width on ResNet-20. For this ablation study, we chose to attack from class 0 to class 9 only. TA is Test accuracy before the attack.}
\label{tab:quan}
\scalebox{0.95}{
\begin{tabular}{@{}ccccccc@{}}
\toprule
Type & \begin{tabular}[c]{@{}c@{}}Attack\\ Success\\ Rate (\%)\end{tabular} & \begin{tabular}[c]{@{}c@{}}Post-Attack \\ Test Accuracy\\ (\%)\end{tabular} & \begin{tabular}[c]{@{}c@{}}\# of\\ Bit-Flips\end{tabular} & \begin{tabular}[c]{@{}c@{}}Attack\\ Success\\ Rate (\%)\end{tabular} & \begin{tabular}[c]{@{}c@{}} Post-Attack \\ Test Accuracy\\ (\%)\end{tabular} & \begin{tabular}[c]{@{}c@{}}\# of\\ Bit-Flips\end{tabular} \\ \midrule
 & \multicolumn{3}{c}{\textbf{8-Bit ( TA: 92.91\%)}} & \multicolumn{3}{c}{\textbf{6-Bit ( TA: 92.41\%)}} \\ \midrule
N-to-1 (I) &100 $\pm$ 0 & 10 $\pm$ 0 & 6.0 $\pm$ 2.2&100 $\pm$ 0 & 10 $\pm$ 0 & 6.2 $\pm$ 0.83 \\
1-to-1 (II) &100 $\pm$ 0 & 68.45 $\pm$ 0.08 & 2.2 $\pm$ 0  & 100 $\pm$ 0 & 68.45 $\pm$ 0.08 & 2 $\pm$ 0  \\
1-to-1 (S) (III) &99.7 $\pm$ 0 & 80.3 $\pm$ 0  & 4.4 $\pm$ 0.54&  100 $\pm$ 0 & 68.53 $\pm$ 0.15 & 1.6 $\pm$ 0.54   \\ \midrule
 & \multicolumn{3}{c}{\textbf{4-Bit (TA: 91.33\%)}} & \multicolumn{3}{c}{\textbf{2-Bit (TA: 90.3\%)}} \\ \midrule
N-to-1 (I) & 100 $\pm$ 0 & 10 $\pm$ 0 & 4.4 $\pm$ 0.55  &  100 $\pm$ 0 & 10 $\pm$ 0 & 37.6 $\pm$ 0.9 \\
1-to-1 (II) &100 $\pm$ 0 & 11.54 $\pm$ 0.08 & 4 $\pm$ 0  &  100 $\pm$ 0 & 76.43 $\pm$ 3.43 & 17.6 $\pm$ 1.7  \\
\multicolumn{1}{l}{1-to-1 (S) (III)} & 98.56 $\pm$ 0.7 & 12.79 $\pm$ 1.88 & 16.4 $\pm$ 9.36 &  97.87 $\pm$ 1.75 & 84.6 $\pm$ 0.91 & 19.6 $\pm$ 6.6 \\ \bottomrule
\end{tabular}}

\end{table*}

\section{Discussion}

\subsection{Evaluation Against Existing Defense} Recently, \cite{he2020defending} proposed Piece-wise Clustering (PC) and Binarization as an effective training scheme to defend against Bit-Flip based un-targeted weight attack \cite{rakin2019bit}. We evaluate our T-BFA against both methods in Table \ref{tab:CVPR20}, showing that T-BFA (e.g., 1-to-1 (S)) still successfully (i.e., higher than 97.0 \% ASR with tens or less \# bit flips) attacks PC and binary network with a cost of around 2 $\times$ and 5 $\times$ more flips, respectively, showing limited resistance improvement, but not significantly. 

In summary, Binarization \cite{he2020defending} is still improving robustness against T-BFA, but not as effective as un-targeted BFA. According to our observation, an N-to-1 attack is a much stronger attack than an un-targeted BFA, meaning it requires 7 times fewer amount of bit-flips to degrade network accuracy to 10 percent in \cref{tab:cmp}. Since T-BFA is more effective than an un-targeted BFA, it also achieves better attack performance against binarization \cite{he2020defending} defense.

\subsection{Effect of Quantization Levels} We perform an ablation study to analyze the effects of weight quantization levels. In Table~\ref{tab:quan}, we show the results of our attack for quantized networks with weights represented by 2,4,6,8 bits. The performance of N-to-1 and 1-to-1 attack is slightly weaker for a low bit-width network (e.g., 2-bit). This observation is in consistent with \cite{he2020defending} where they demonstrate network capacity (e.g., model size) plays a key role in increasing the difficulty of BFA attack. In summary, network with high quantization level resists (i.e., poor test accuracy after attack) 1-to-1 (S) attack better. While, network with lower bit resolution (e.g., in \cref{tab:quan}) has a better resistance against N-to-1 attack.

\subsection{Layer-wise sensitivity.} We also observe that the most vulnerable or sensitive layer under the T-BFA attack is the last classification layer. In the case of 1-to-1 (s) attack, it is interesting to observe that 100\% of all the identified vulnerable bits are in the last layer for both ResNet-20 and VGG-11 models. For the N-to-1 attack, more than 90\% of bit-flips are in the last classification layer. This study leads to the question: \textit{Can we defend T-BFA by securing the critical last layer for classification?} To answer this question, we assume the entire last layer is protected (i.e. no bit-flip is allowed) and run the T-BFA again. This is motivated by prior work that secures the entire last layer in a protected enclave of computer processor, such as Intel SGX \cite{tramer2018slalom}, as an effective privacy protection method. Unfortunately, all three versions of T-BFA still succeed with a cost of limited additional number (all less than 30) of bit flips. Thus, this scheme helps slightly improve the resistance, but not significantly.

\begin{table}[ht]

\centering
\caption{Summary of possible directions to improve resistance against T-BFA.}
\label{tabel:checklist}
\resizebox{0.49\textwidth}{!}{%
\begin{tabular}{@{}lcccc@{}}
\toprule
\multicolumn{1}{c}{Possible Defense Directions} & I & II & III  \\ \midrule
1. Perform Weight Clustering (i.e., Binarization \& PC) & \checkmark& \checkmark& \checkmark \\
2. Increase Network Capacity (e.g., larger size/ high bit-width) & \checkmark & \checkmark &  \\
3. Decrease Network Capacity (e.g., smaller network) &  &  & \checkmark \\
4. Securing critical layers (e.g., Classification layer)  & \checkmark& \checkmark& \checkmark \\

\end{tabular}%
}

\end{table}

\subsection{Summary of Potential Defenses}
Based on above discussion and our summarized take-aways, we list several directions we have explored to improve DNN model resistance against different types of T-BFA in Table \ref{tabel:checklist}. 
In our experiments, those methods only help improve DNN model resistance in a limited degree. However, none of them could significantly improve Robustness. For example, the largest bit-flip \# for any type of T-BFA to succeed on CIFAR-10 is $36$ when attacking binary network in \cref{tab:CVPR20}, which is still a practical number in real-computer memory fault injection as discussed in \cite{yao2020deephammer,cojocar19exploiting}.

\section{Conclusion}
In this work, as far as we know, we are the first to propose three targeted adversarial weight attack schemes, i.e. N-to-1, 1-to-1 and 1-to-1(stealthy), which severely degrade the classification performance of quantized DNNs. Our T-BFA is based on an iterative class-dependent bit ranking algorithm. Extensive experiments have been conducted to prove the efficacy of our proposed T-BFA in different DNN architectures on CIFAR10 and Imagenet datasets. Moreover, we also demonstrate our T-BFA in a real-computer running DNNs. In the end, we provide several possible analysis and directions to construct robust DNN models against T-BFA. 

\appendices

\ifCLASSOPTIONcompsoc

\ifCLASSOPTIONcaptionsoff
  \newpage
\fi



%

\bibliographystyle{IEEEtran}
\scriptsize
\bibliography{IEEEabrv,./reference}

\begin{thebibliography}{10}
\providecommand{\url}[1]{#1}
\csname url@samestyle\endcsname
\providecommand{\newblock}{\relax}
\providecommand{\bibinfo}[2]{#2}
\providecommand{\BIBentrySTDinterwordspacing}{\spaceskip=0pt\relax}
\providecommand{\BIBentryALTinterwordstretchfactor}{4}
\providecommand{\BIBentryALTinterwordspacing}{\spaceskip=\fontdimen2\font plus
\BIBentryALTinterwordstretchfactor\fontdimen3\font minus
  \fontdimen4\font\relax}
\providecommand{\BIBforeignlanguage}[2]{{%
\expandafter\ifx\csname l@#1\endcsname\relax
\typeout{** WARNING: IEEEtran.bst: No hyphenation pattern has been}%
\typeout{** loaded for the language `#1'. Using the pattern for}%
\typeout{** the default language instead.}%
\else
\language=\csname l@#1\endcsname
\fi
#2}}
\providecommand{\BIBdecl}{\relax}
\BIBdecl

\bibitem{krizhevsky2010cifar}
A.~Krizhevsky, V.~Nair, and G.~Hinton, ``Cifar-10 (canadian institute for
  advanced research),'' \emph{URL http://www. cs. toronto. edu/kriz/cifar.
  html}, 2010.

\bibitem{sun2019evolving}
Y.~Sun, B.~Xue, M.~Zhang, and G.~G. Yen, ``Evolving deep convolutional neural
  networks for image classification,'' \emph{IEEE Transactions on Evolutionary
  Computation}, 2019.

\bibitem{hinton2012deep}
G.~Hinton, L.~Deng, D.~Yu, G.~E. Dahl, A.-r. Mohamed, N.~Jaitly, A.~Senior,
  V.~Vanhoucke, P.~Nguyen, T.~N. Sainath \emph{et~al.}, ``Deep neural networks
  for acoustic modeling in speech recognition: The shared views of four
  research groups,'' \emph{IEEE Signal Processing Magazine}, vol.~29, no.~6,
  pp. 82--97, 2012.

\bibitem{haque2020experimental}
M.~A. Haque, A.~Verma, J.~S.~R. Alex, and N.~Venkatesan, ``Experimental
  evaluation of cnn architecture for speech recognition,'' in \emph{First
  International Conference on Sustainable Technologies for Computational
  Intelligence}.\hskip 1em plus 0.5em minus 0.4em\relax Springer, 2020, pp.
  507--514.

\bibitem{luong2015deep}
M.-T. Luong, M.~Kayser, and C.~D. Manning, ``Deep neural language models for
  machine translation,'' in \emph{Proceedings of the Nineteenth Conference on
  Computational Natural Language Learning}, 2015, pp. 305--309.

\bibitem{lu2020research}
Y.~Lu, X.~Xiong, W.~Zhang, J.~Liu, and R.~Zhao, ``Research on classification
  and similarity of patent citation based on deep learning,''
  \emph{Scientometrics}, pp. 1--27, 2020.

\bibitem{madry2018towards}
\BIBentryALTinterwordspacing
A.~Madry, A.~Makelov, L.~Schmidt, D.~Tsipras, and A.~Vladu, ``Towards deep
  learning models resistant to adversarial attacks,'' in \emph{International
  Conference on Learning Representations}, 2018. [Online]. Available:
  \url{https://openreview.net/forum?id=rJzIBfZAb}
\BIBentrySTDinterwordspacing

\bibitem{goodfellow2014explaining}
I.~J. Goodfellow, J.~Shlens, and C.~Szegedy, ``Explaining and harnessing
  adversarial examples,'' \emph{ICLR}, 2015.

\bibitem{yan2020cache}
M.~Yan, C.~W. Fletcher, and J.~Torrellas, ``Cache telepathy: Leveraging shared
  resource attacks to learn $\{$DNN$\}$ architectures,'' in \emph{29th
  $\{$USENIX$\}$ Security Symposium ($\{$USENIX$\}$ Security 20)}, 2020, pp.
  2003--2020.

\bibitem{xiang2020open}
Y.~Xiang, Z.~Chen, Z.~Chen, Z.~Fang, H.~Hao, J.~Chen, Y.~Liu, Z.~Wu, Q.~Xuan,
  and X.~Yang, ``Open dnn box by power side-channel attack,'' \emph{IEEE
  Transactions on Circuits and Systems II: Express Briefs}, 2020.

\bibitem{yudeepem}
H.~Yu, H.~Ma, K.~Yang, Y.~Zhao, and Y.~Jin, ``Deepem: Deep neural networks
  model recovery through em side-channel information leakage,'' 2020.

\bibitem{das2019x}
D.~Das, A.~Golder, J.~Danial, S.~Ghosh, A.~Raychowdhury, and S.~Sen,
  ``X-deepsca: Cross-device deep learning side channel attack,'' in
  \emph{Proceedings of the 56th Annual Design Automation Conference 2019},
  2019, pp. 1--6.

\bibitem{roscian2013fault}
C.~Roscian, A.~Sarafianos, J.-M. Dutertre, and A.~Tria, ``Fault model analysis
  of laser-induced faults in sram memory cells,'' in \emph{2013 Workshop on
  Fault Diagnosis and Tolerance in Cryptography}.\hskip 1em plus 0.5em minus
  0.4em\relax IEEE, 2013, pp. 89--98.

\bibitem{kim2014flipping}
Y.~Kim, R.~Daly, J.~Kim, C.~Fallin, J.~H. Lee, D.~Lee, C.~Wilkerson, K.~Lai,
  and O.~Mutlu, ``Flipping bits in memory without accessing them: An
  experimental study of dram disturbance errors,'' in \emph{ACM SIGARCH
  Computer Architecture News}, vol.~42, no.~3.\hskip 1em plus 0.5em minus
  0.4em\relax IEEE Press, 2014, pp. 361--372.

\bibitem{razavi2016flip}
K.~Razavi, B.~Gras, E.~Bosman, B.~Preneel, C.~Giuffrida, and H.~Bos, ``Flip
  feng shui: Hammering a needle in the software stack,'' in \emph{25th
  $\{$USENIX$\}$ Security Symposium ($\{$USENIX$\}$ Security 16)}, 2016, pp.
  1--18.

\bibitem{rakin2019bit}
A.~S. Rakin, Z.~He, and D.~Fan, ``Bit-flip attack: Crushing neural network with
  progressive bit search,'' in \emph{Proceedings of the IEEE International
  Conference on Computer Vision (ICCV)}, 2019, pp. 1211--1220.

\bibitem{liu2017fault}
Y.~Liu, L.~Wei, B.~Luo, and Q.~Xu, ``Fault injection attack on deep neural
  network,'' in \emph{2017 IEEE/ACM International Conference on Computer-Aided
  Design (ICCAD)}.\hskip 1em plus 0.5em minus 0.4em\relax IEEE, 2017, pp.
  131--138.

\bibitem{hong2019terminal}
S.~Hong, P.~Frigo, Y.~Kaya, C.~Giuffrida, and T.~Dumitraș, ``Terminal brain
  damage: Exposing the graceless degradation in deep neural networks under
  hardware fault attacks,'' in \emph{28th $\{$USENIX$\}$ Security Symposium
  ($\{$USENIX$\}$ Security 19)}, 2019, pp. 497--514.

\bibitem{yao2020deephammer}
F.~Yao, A.~S. Rakin, and D.~Fan, ``Deephammer: Depleting the intelligence of
  deep neural networks through targeted chain of bit flips,'' \emph{29th
  $\{$USENIX$\}$ Security Symposium ($\{$USENIX$\}$ Security)}, 2020.

\bibitem{chen2017zoo}
P.-Y. Chen, H.~Zhang, Y.~Sharma, J.~Yi, and C.-J. Hsieh, ``Zoo: Zeroth order
  optimization based black-box attacks to deep neural networks without training
  substitute models,'' in \emph{Proceedings of the 10th ACM Workshop on
  Artificial Intelligence and Security}.\hskip 1em plus 0.5em minus 0.4em\relax
  ACM, 2017, pp. 15--26.

\bibitem{carlini2017towards}
N.~Carlini and D.~Wagner, ``Towards evaluating the robustness of neural
  networks,'' in \emph{Security and Privacy (SP), 2017 IEEE Symposium
  on}.\hskip 1em plus 0.5em minus 0.4em\relax IEEE, 2017, pp. 39--57.

\bibitem{zhao2019fault}
P.~Zhao, S.~Wang, C.~Gongye, Y.~Wang, Y.~Fei, and X.~Lin, ``Fault sneaking
  attack: A stealthy framework for misleading deep neural networks,'' in
  \emph{2019 56th ACM/IEEE Design Automation Conference (DAC)}.\hskip 1em plus
  0.5em minus 0.4em\relax IEEE, 2019, pp. 1--6.

\bibitem{jouppi2017datacenter}
N.~P. Jouppi, C.~Young, N.~Patil, D.~Patterson, G.~Agrawal, R.~Bajwa, S.~Bates,
  S.~Bhatia, N.~Boden, A.~Borchers \emph{et~al.}, ``In-datacenter performance
  analysis of a tensor processing unit,'' in \emph{Proceedings of the 44th
  Annual International Symposium on Computer Architecture}, 2017, pp. 1--12.

\bibitem{agoyan2010flip}
M.~Agoyan, J.-M. Dutertre, A.-P. Mirbaha, D.~Naccache, A.-L. Ribotta, and
  A.~Tria, ``How to flip a bit?'' in \emph{2010 IEEE 16th International On-Line
  Testing Symposium}.\hskip 1em plus 0.5em minus 0.4em\relax IEEE, 2010, pp.
  235--239.

\bibitem{rakin2019tbt}
A.~S. Rakin, Z.~He, and D.~Fan, ``Tbt: Targeted neural network attack with bit
  trojan,'' \emph{Proceedings of the IEEE Conference on Computer Vision and
  Pattern Recognition}, 2020.

\bibitem{bengio2013estimating}
Y.~Bengio, N.~L{\'e}onard, and A.~Courville, ``Estimating or propagating
  gradients through stochastic neurons for conditional computation,''
  \emph{arXiv preprint arXiv:1308.3432}, 2013.

\bibitem{gu2019badnets}
T.~Gu, K.~Liu, B.~Dolan-Gavitt, and S.~Garg, ``Badnets: Evaluating backdooring
  attacks on deep neural networks,'' \emph{IEEE Access}, vol.~7, pp.
  47\,230--47\,244, 2019.

\bibitem{Trojannn}
Y.~Liu, S.~Ma, Y.~Aafer, W.-C. Lee, J.~Zhai, W.~Wang, and X.~Zhang, ``Trojaning
  attack on neural networks,'' in \emph{25nd Annual Network and Distributed
  System Security Symposium, {NDSS} 2018, San Diego, California, USA, February
  18-221, 2018}.\hskip 1em plus 0.5em minus 0.4em\relax The Internet Society,
  2018.

\bibitem{krizhevsky2012imagenet}
A.~Krizhevsky, I.~Sutskever, and G.~E. Hinton, ``Imagenet classification with
  deep convolutional neural networks,'' in \emph{Advances in neural information
  processing systems}, 2012, pp. 1097--1105.

\bibitem{he2016deep}
K.~He, X.~Zhang, S.~Ren, and J.~Sun, ``Deep residual learning for image
  recognition,'' in \emph{Proceedings of the IEEE conference on computer vision
  and pattern recognition}, 2016, pp. 770--778.

\bibitem{simonyan2014very}
K.~Simonyan and A.~Zisserman, ``Very deep convolutional networks for
  large-scale image recognition,'' \emph{arXiv preprint arXiv:1409.1556}, 2014.

\bibitem{he2020defending}
Z.~He, A.~S. Rakin, J.~Li, C.~Chakrabarti, and D.~Fan, ``Defending and
  harnessing the bit-flip based adversarial weight attack,'' in
  \emph{Proceedings of the IEEE/CVF Conference on Computer Vision and Pattern
  Recognition}, 2020, pp. 14\,095--14\,103.

\bibitem{sandler2018mobilenetv2}
M.~Sandler, A.~Howard, M.~Zhu, A.~Zhmoginov, and L.-C. Chen, ``Mobilenetv2:
  Inverted residuals and linear bottlenecks,'' in \emph{Proceedings of the IEEE
  Conference on Computer Vision and Pattern Recognition}, 2018, pp. 4510--4520.

\bibitem{pessl2016drama}
P.~Pessl, D.~Gruss, C.~Maurice, M.~Schwarz, and S.~Mangard, ``{DRAMA}:
  Exploiting {DRAM} addressing for cross-cpu attacks,'' in \emph{{USENIX}
  Security Symposium}, 2016, pp. 565--581.

\bibitem{gruss2018another}
D.~Gruss, M.~Lipp, M.~Schwarz, D.~Genkin, J.~Juffinger, S.~O'Connell,
  W.~Schoechl, and Y.~Yarom, ``Another flip in the wall of rowhammer
  defenses,'' in \emph{2018 IEEE Symposium on Security and Privacy (SP)}.\hskip
  1em plus 0.5em minus 0.4em\relax IEEE, 2018, pp. 245--261.

\bibitem{tramer2018slalom}
\BIBentryALTinterwordspacing
F.~Tramer and D.~Boneh, ``Slalom: Fast, verifiable and private execution of
  neural networks in trusted hardware,'' in \emph{International Conference on
  Learning Representations}, 2019. [Online]. Available:
  \url{https://openreview.net/forum?id=rJVorjCcKQ}
\BIBentrySTDinterwordspacing

\bibitem{cojocar19exploiting}
L.~Cojocar, K.~Razavi, C.~Giuffrida, and H.~Bos, ``Exploiting correcting codes:
  On the effectiveness of ecc memory against rowhammer attacks.''

\end{thebibliography}

%




\end{document}